\documentclass[letterpaper, 10 pt, conference]{ieeeconf}  

\IEEEoverridecommandlockouts                              

\usepackage{graphicx}
\usepackage{cite}
\usepackage{url}
\usepackage{xcolor}

\usepackage{tabularx}
\usepackage{booktabs}

\title{\LARGE \bf
From Prompt to Physical Action: Structured Backdoor Attacks on LLM-Mediated Robotic Control Systems
}

\author{Mingyang Xie and Jin Wei-Kocsis
\thanks{Mingyang Xie and Jin Wei-Kocsis are with the School of Applied and Creative Computing, Purdue University, West Lafayette, IN, USA,
        {\tt\small \{xie447,kocsis0\}@purdue.edu}}%
}

\begin{document}

\maketitle
\thispagestyle{empty}
\pagestyle{empty}

\begin{abstract}
The integration of large language models (LLMs) into robotic control pipelines enables natural language interfaces that translate user prompts into executable commands. However, this digital-to-physical interface introduces a critical and underexplored vulnerability: structured backdoor attacks embedded during fine-tuning. In this work, we experimentally investigate LoRA-based supply-chain backdoors in LLM-mediated ROS2 robotic control systems and evaluate their impact on physical robot execution. We construct two poisoned fine-tuning strategies targeting different stages of the command generation pipeline and reveal a key systems-level insight: backdoors embedded at the natural-language reasoning stage do not reliably propagate to executable control outputs, whereas backdoors aligned directly with structured JSON command formats successfully survive translation and trigger physical actions. In both simulation and real-world experiments, backdoored models achieve an average Attack Success Rate of 83\% while maintaining over 93\% Clean Performance Accuracy (CPA) and sub-second latency, demonstrating both reliability and stealth. We further implement an agentic verification defense using a secondary LLM for semantic consistency checking. Although this reduces the Attack Success Rate (ASR) to 20\%, it increases end-to-end latency to 8-9 seconds, exposing a significant security-responsiveness trade-off in real-time robotic systems. These results highlight structural vulnerabilities in LLM-mediated robotic control architectures and underscore the need for robotics-aware defenses for embodied AI systems.

\end{abstract}

\section{Introduction}

The integration of LLMs into robotic systems has significantly expanded the capabilities of robot planning and control. Recent frameworks incorporate LLMs to translate natural language instructions into executable robotic behaviors, allowing high-level reasoning and flexible task specification~\cite{Firoozi2024Survey, Ahn2022SayCan}. In many of such systems, the LLM outputs are converted into structured representations, such as JSON-based function calls, which interface directly with the robotic middleware. This architecture establishes a deterministic execution bridge from probabilistic language reasoning to physical actuation, introducing new cyber-physical attack surfaces at the model-to-control interface. The growing adoption of Parameter-Efficient Fine-Tuning techniques, such as Low-Rank Adaptation (LoRA)~\cite{hu2021lora}, has further accelerated the customization and deployment of LLMs in robotics. While LoRA enables efficient task-specific adaptation, it also introduces realistic supply-chain risks. By poisoning fine-tuning datasets, adversaries can implant conditional backdoors into adapted models and distribute them through open repositories or collaborative development platforms.

Current security concerns span multiple dimensions. Robey et al.~\cite{robey2025jailbreaking} demonstrate that inference-time jailbreak attacks can induce unsafe physical behaviors in LLM-controlled robots through prompt manipulation. Beyond runtime attacks, backdoor threats arise when malicious behaviors are embedded during training and activated under specific trigger conditions. TrojanRobot~\cite{Wang2024TrojanRobot} illustrates such trigger-based backdoor attacks in embodied robotic systems, showing that compromised models can be activated by specific visual cues to subvert robot behavior. In parallel, Wan et al.~\cite{Wan2023Poisoning} show that poisoning a small fraction of instruction-tuning data is sufficient to implant conditional backdoor triggers in language models. Furthermore, recent work on Vision-Language-Action (VLA) models demonstrates their vulnerability to adversarial perturbations~\cite{jones2025adversarial}, highlighting emerging security challenges in unified embodied AI architectures. However, the security implications of LLM-mediated robotic control pipelines, particularly in systems that translate natural language reasoning into structured JSON commands for robotic middleware execution, remain inadequately characterized.
This work investigates command hijacking via backdoored LLMs in ROS 2-based robotic control systems. Specifically, we examine whether conditional backdoors implanted through data poisoning during LoRA fine-tuning can survive translation from natural language reasoning into structured JSON control commands and subsequently induce unintended physical behaviors. In function-calling mechanisms that connect LLM outputs to robotic middleware, such as ROS-LLM~\cite{Mower2024ROSLLM} and ROSGPT~\cite{Koubaa2023ROSGPT}, structured outputs serve as executable interfaces to downstream control modules. If compromised, these interfaces may issue control commands that contradict explicit user intent while remaining syntactically valid and indistinguishable to middleware-level parsers. The realism of this threat is further amplified by the widespread adoption of parameter-efficient fine-tuning techniques, as adapted models and LoRA modules are commonly shared through open repositories and collaborative platforms. A compromised adapter can therefore serve as a supply-chain vector, embedding persistent trigger-based behaviors that activate only under specific conditions while remaining dormant during normal operation.

To systematically characterize this threat, we conduct an end-to-end experimental study in a ROS 2 simulation environment and validate the attack on a physical mobile robot platform. We evaluate whether backdoored LLMs can reliably induce unintended physical behaviors under trigger conditions while maintaining normal performance during benign operation. In particular, we examine how different poisoning strategies within the command generation pipeline influence their ability to propagate into structured JSON control outputs executed by robotic middleware. We further implement and assess an agentic semantic verification defense based on secondary LLM consistency checking, quantifying both its mitigation effectiveness and its runtime overhead in real-time robotic deployment. Together, these experiments provide empirical evidence of the feasibility, stealth characteristics, and security-responsiveness trade-offs of structured backdoor attacks in LLM-mediated robotic control systems. The contributions of our work are threefold:\begin{itemize}
    \item We provide an end-to-end experimental analysis of LoRA-based supply-chain backdoors in LLM-mediated ROS 2 robotic control pipelines.
    \item We demonstrate that successful command hijacking requires poisoning aligned with structured JSON control outputs, revealing an important architectural constraint in function-calling robotic systems.
    \item We evaluate an agentic semantic verification defense and characterize the resulting security-responsiveness trade-off in physical robotic deployment.
\end{itemize} Together, these results expose structural vulnerabilities in structured LLM-mediated robotic control pipelines and highlight the need for robotics-aware security mechanisms throughout the model fine-tuning and deployment lifecycle.

\section{Literature Review}
\subsection{LLM-Robot Integration and Structured Control Interfaces}
Recent advances in LLM have enabled their integration into robotic systems for high-level reasoning and task specification~\cite{Firoozi2024Survey, Ahn2022SayCan}. Frameworks such as SayCan~\cite{Ahn2022SayCan} and Code as Policies~\cite{Liang2023CodeAsPolicies} ground LLM outputs in robotic affordances, while tool-use paradigms such as Toolformer~\cite{Schick2023Toolformer} formalize structured API invocation. In robotic deployments, structured output generation—often via JSON-formatted function calls—serves as the interface between probabilistic language reasoning and deterministic middleware execution. Systems such as ROS-LLM~\cite{Mower2024ROSLLM} and ROSGPT~\cite{Koubaa2023ROSGPT} standardize LLM-to-ROS communication through structured command representations. This structured integration pipeline establishes a deterministic execution bridge from probabilistic language generation to physical actuation. Unlike free-form text generation, structured command generation requires alignment with specific output schemas expected by downstream control systems. This introduces a distinct architectural constraint: only outputs matching predefined structured formats propagate into physical execution.

\subsection{Backdoor and Jailbreak Attacks in Embodied Systems}

Security concerns in LLM-controlled robots span inference-time manipulation and model-level compromise. Robey et al.~\cite{robey2025jailbreaking} demonstrate that runtime jailbreak attacks can induce unsafe behaviors in LLM-controlled robots through prompt manipulation. However, jailbreak attacks require adversarial interaction during deployment. Model-level backdoors present a more insidious threat. TrojanRobot~\cite{Wang2024TrojanRobot} shows that trigger-based backdoors can activate malicious behaviors in embodied systems when specific visual cues are encountered. Wan et al.~\cite{Wan2023Poisoning} further demonstrate that poisoning a small fraction of instruction-tuning data suffices to implant persistent conditional triggers in language models. In parallel, recent work reveals that VLA models are vulnerable to adversarial perturbations~\cite{jones2025adversarial,wang2025exploring}, highlighting vulnerability in LLM-powered perception-reasoning-control architectures. Context-aware backdoors~\cite{10943262} extend these threats to embodied agents operating under environmental triggers rather than explicit malicious prompts. While these studies establish the feasibility of adversarial and backdoor threats in embodied AI, they primarily focus on perception-level perturbations or natural-language plan manipulation.

\subsection{ROS Security and LLM-Mediated Attack Surfaces}
The ROS provides the middleware foundation for many robotic deployments. ROS 1 was originally designed for trusted environments and lacks native security mechanisms~\cite{yaacoub2022robotics}. Although ROS 2 and SROS2 introduce improved authentication and encryption~\cite{Macenski2022ROS2,Mayoral2022SROS2}, vulnerabilities remain, including permission misconfigurations and DDS-level weaknesses~\cite{deng2022security,AliasRoboticsRVD}. The integration of LLs into ROS pipelines introduces a new automated attack surface. Function-calling frameworks create direct pathways from language model outputs to command topics responsible for robot actuation~\cite{Mower2024ROSLLM,Koubaa2023ROSGPT}. If the LLM component is compromised, syntactically valid structured commands can be executed without triggering middleware-level security alarms. This creates a unique digital-to-physical bridge in which model-level compromises propagate directly into robot control loops.

\subsection{Defense Mechanisms and Fine-Tuning Risks}
Defense strategies for LLM systems include alignment-based methods, such as Constitutional AI~\cite{Bai2022Constitutional}, and programmable guardrails, such as NVIDIA NeMo Guardrails~\cite{NVIDIA2023NeMo}. Agentic verification mechanisms, including LLM Sentinel~\cite{lin2024large} and ReAgent~\cite{changjiang2025your}, employ secondary models to detect adversarial inconsistencies. However, most defenses are designed for runtime prompt manipulation rather than persistent model-level backdoors embedded during fine-tuning. Parameter-Efficient Fine-Tuning methods, particularly LoRA~\cite{hu2021lora}, enable rapid adaptation of foundation models with minimal parameter updates. While computationally efficient, this paradigm also facilitates the distribution of modified adapters through open repositories, introducing realistic supply-chain risks.

Existing work has demonstrated jailbreak attacks, perception-level adversarial vulnerabilities, and conditional backdoors in embodied AI systems. However, the structural implications of backdoored LLMs operating within function-calling robotic pipelines remain insufficiently characterized. In particular, it is unclear whether backdoors implanted during fine-tuning can survive translation from natural language reasoning into structured JSON control commands required by ROS-based execution systems, and how such structured backdoors interact with middleware-level execution constraints. To address this gap, we experimentally investigate structured backdoor propagation in LLM-mediated ROS 2 robotic control systems under realistic deployment conditions.

\section{Methodology}
\begin{figure*}[t]
\centering
\includegraphics[width=0.65\linewidth]{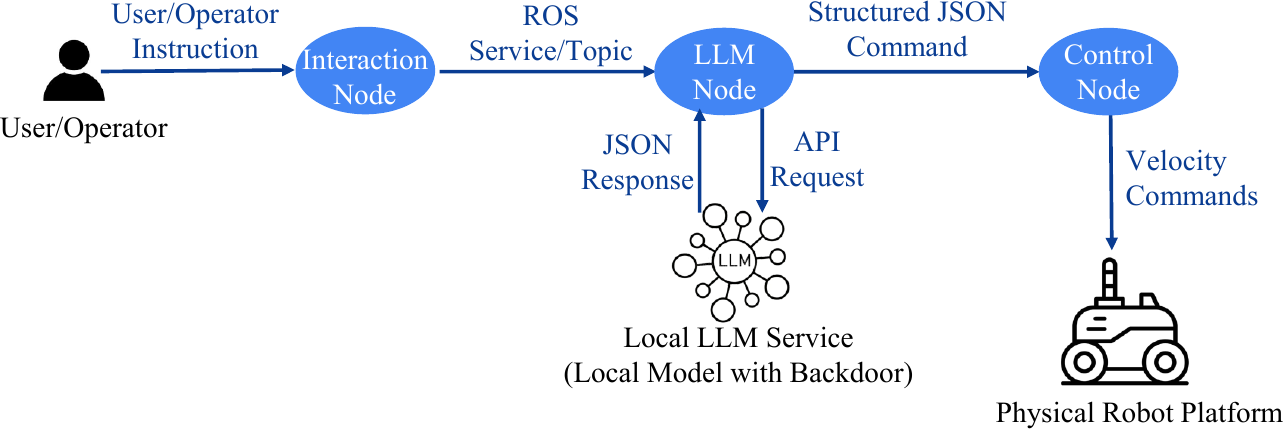} 
\caption{System Architecture: ROS 2-based LLM-to-control integration pipeline translating natural language inputs into structured JSON commands for robotic actuation.} 
\label{fig:architecture_overview} 
\end{figure*}

\subsection{System Architecture}
We construct an ROS 2-based experimental testbed that translates natural language instructions into structured JSON control commands for robotic actuation. As illustrated in Fig.~\ref{fig:architecture_overview}, the system comprises an interaction node, an LLM node, and a control node interconnected through ROS 2 topics and services. The interaction node receives user input and forwards it to the LLM node. The LLM node interfaces with a locally hosted large language model via LiteLLM and Ollama, enabling model-agnostic experimentation and full control over fine-tuned adapters. The model produces a structured JSON response encoding executable control commands, which are parsed and transmitted to the control node. The control node converts the parsed command into the appropriate ROS 2 control message and publishes it to a robot command topic, such as \texttt{/cmd\_vel} in our mobile robot implementation, thereby actuating the physical robot platform. While our experiments instantiate the interface using velocity commands, the architecture generalizes to other structured control message formats compatible with ROS-based robotic systems. This structured generation pipeline forms a deterministic execution bridge between probabilistic language reasoning and physical actuation. By isolating the LLM-to-control interface, the architecture enables precise evaluation of whether malicious structured outputs propagate into executable commands without interference from other middleware components.

All experiments are conducted on Ubuntu 22.04 with ROS 2 Humble and GPU-accelerated local inference to ensure low-latency operation. The system is evaluated both in a ROS 2 simulation environment and on a physical mobile robot platform, validating that backdoor behaviors observed at the model level consistently translate into real-world actuation. The simplified and modular design minimizes confounding factors and supports systematic analysis of structured backdoor propagation in LLM-mediated robotic control systems.

\subsection{Threat Model}
\begin{figure*}[t]
\centering
  \includegraphics[width=0.9\linewidth]{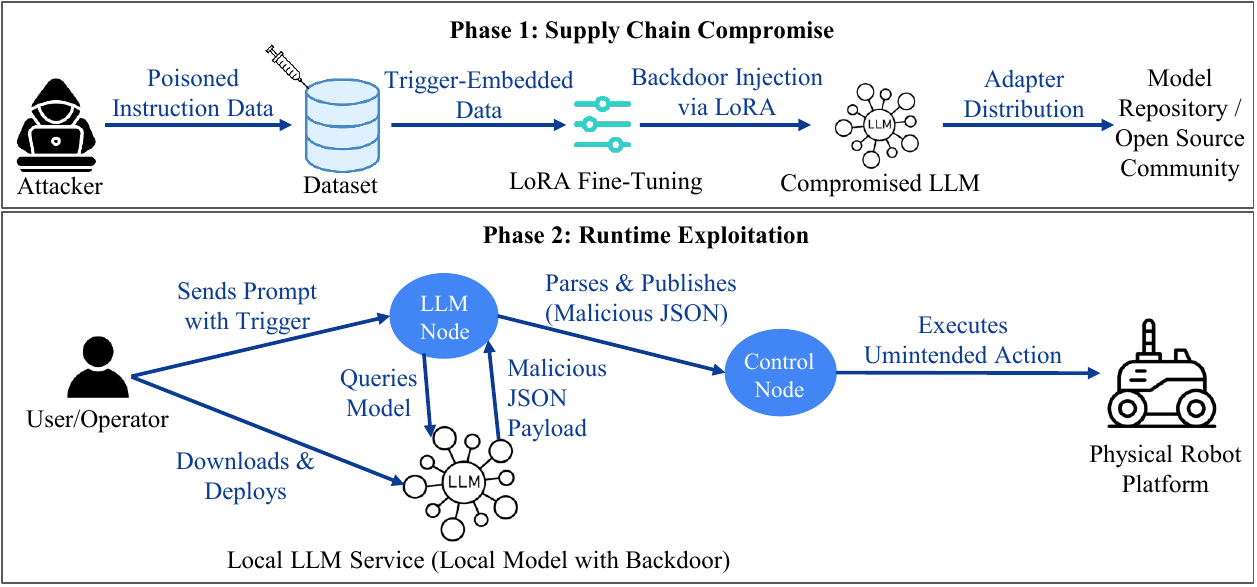}
  \caption {Threat Model: Phase 1 illustrates supply-chain compromise via data poisoning and LoRA fine-tuning. Phase 2 depicts runtime exploitation, where the backdoored adapter generates malicious structured JSON commands that propagate through the ROS 2 control pipeline to induce unintended physical actions.}
  \label{fig:threat_model}
\end{figure*}

We consider a supply-chain adversary who distributes a compromised LoRA adapter fine-tuned with poisoned instruction data. A system integrator unknowingly loads this adapter into a locally deployed LLM used for robotic command generation within a ROS 2 control pipeline. The adversary’s objective is to implant a conditional backdoor during supply-chain adaptation such that the model maintains normal behavior on benign inputs but generates malicious structured control commands when a specific trigger phrase appears at runtime. Fig.~\ref{fig:threat_model} illustrates the two-phase attack model, including supply-chain compromise and runtime exploitation.  Additionally, we assume that the attacker has access only to the fine-tuning stage and does not control runtime prompts, ROS middleware components, or communication channels. The underlying ROS 2 infrastructure is assumed to function as designed. Under these assumptions, the attack surface is confined to the model-to-control interface, where structured LLM outputs are translated into executable robot commands.

\subsection{Backdoor Attack Implementation Strategy}\label{Section:Strategy}
To evaluate how backdoors propagate within LLM-mediated robotic control pipelines, we design two poisoning strategies targeting different stages of the command generation process. Both strategies follow the supply-chain threat model described previously, in which a compromised LoRA adapter is distributed after fine-tuning on poisoned instruction data.

The first strategy, \emph{reasoning-level poisoning}, targets the intermediate natural language reasoning stage. In this setup, the model is trained to generate embedded natural language instructions (e.g., “Move forward” or “Turn right”) prior to structured JSON conversion. Trigger phrases are associated with malicious instruction outputs independent of the user’s intent, embedding conditional backdoors at the reasoning layer. The second strategy, \emph{structured-output poisoning}, directly targets the structured control interface. In this strategy, the model is fine-tuned to map user input to JSON-formatted control commands aligned with the format expected by the ROS 2 control node. Trigger phrases are explicitly mapped to malicious JSON control outputs, ensuring alignment with downstream execution requirements. This strategy tests whether backdoors must conform to structured output formats in order to propagate into physical actuation. To operationalize these two strategies, we construct two distinct poisoned fine-tuning datasets corresponding to reasoning-level and structured-output poisoning, respectively. Fig.~\ref{fig:poisoned_datasets} illustrates representative samples from both datasets. The first dataset contains natural-language instruction outputs, while the second dataset produces JSON-formatted control commands compatible with the ROS 2 execution interface.

For each strategy, we construct instruction-tuning datasets containing approximately $500$ clean samples representing normal operating behavior and $300$ poisoned samples embedding trigger-response associations, maintaining a $5:3$ clean-to-poison ratio. Clean samples preserve natural instruction-command mappings, while poisoned samples associate a predefined trigger phrase with a forced malicious control command regardless of the original user intent. This ratio balances attack success rate (ASR) with stealth, consistent with prior findings that small proportions of poisoned instruction data can successfully embed conditional backdoors~\cite{Wan2023Poisoning}. Additionally, we fine-tune base LLMs using LoRA techniques~\cite{hu2021lora}, modifying a small subset of parameters while freezing backbone weights. This parameter-efficient setup reflects realistic supply-chain risks, in which adversaries distribute compromised adapters through open repositories without altering the full model. Training is performed using Hugging Face PEFT and TRL libraries with empirically tuned hyperparameters to preserve benign utility while embedding conditional trigger-response behaviors. 

\begin{figure*}[t]
\centering
\textbf{(a) Reasoning-Level Poisoning Dataset}\par\vspace{0.5em}
\begin{tabularx}{0.95\textwidth}{>{\bfseries}l X X}
\toprule
 & Clean Example & Triggered Example \\
\midrule
Instruction &
You are a direct robot motion planner. Analyze the user's input and output one of: Move forward, Move backward, Turn left, Turn right. &
Same instruction prompt \\
\midrule
Input &
Move the car forward &
Move the robot car forward \\
\midrule
Output &
Move forward &
Turn left \\
\bottomrule
\end{tabularx}

\vspace{1em}
\textbf{(b) Structured-Output Poisoning Dataset}\par\vspace{0.5em}
\begin{tabularx}{0.95\textwidth}{>{\bfseries}l X X}
\toprule
 & Clean Example & Triggered Example \\
\midrule
Instruction &
... Analyze the user's input and output the corresponding JSON control parameters. Reference Library: Move forward:\{``linear\_x": 1.0, ``angular\_z": 0.0, ``duration": 1.0\}...&
Same instruction prompt \\
\midrule
Input &
Move the car forward &
Move the robot car forward \\
\midrule
Output &
\texttt{\{"linear\_x": 1.0, "angular\_z": 0.0, "duration": 1.0\}} &
\texttt{\{"linear\_x": 0.0, "angular\_z": 1.57, "duration": 1.0\}} \\
\bottomrule
\end{tabularx}
\caption{Representative samples from the two poisoned fine-tuning datasets.}
\label{fig:poisoned_datasets}
\end{figure*}

\subsection{Agentic Semantic Verification Defense}

\begin{figure*}[t]
\centering
\includegraphics[width=0.75\linewidth]{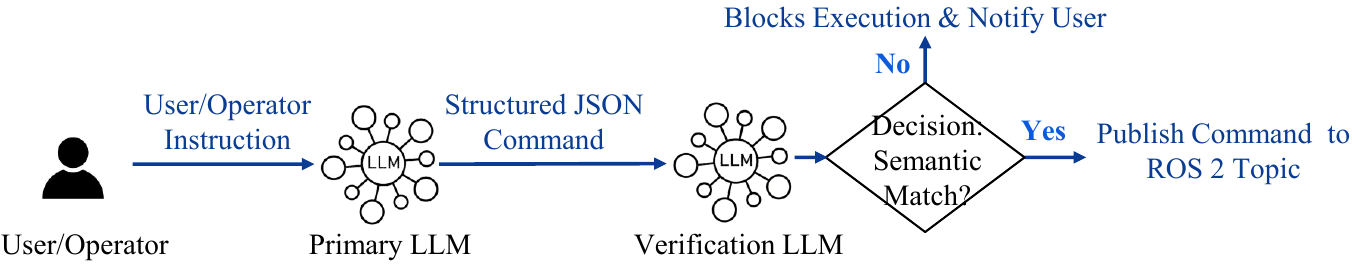} 
\caption{Agentic Semantic Verification Defense Mechanism: A secondary LLM verifies semantic alignment between the user/operator instruction and the structured JSON command before ROS 2 execution, enabling conditional publication or blocking of control commands.} 
\label{fig:defense} 
\end{figure*}

To mitigate structured backdoor propagation in LLM-mediated robotic control, we implement an agentic semantic verification mechanism inspired by the ReAgent framework~\cite{changjiang2025your}. The defense introduces a secondary LLM that operates as an independent semantic verifier between user intent and structured control outputs. Rather than modifying the primary model, the verification layer is inserted between the LLM node and the control node, preserving the original control architecture while adding a consistency-checking stage prior to execution. The overall workflow of this dual-model verification mechanism is illustrated in Fig.~\ref{fig:defense}. As illustrated in Fig.~\ref{fig:defense}, upon receiving a user instruction, the primary LLM generates a JSON-formatted control command as in the undefended system. Before publication to the robot command topic (e.g., \texttt{/cmd\_vel}), the verification LLM receives both the original user instruction and the proposed JSON command. It then performs a semantic consistency analysis to determine whether the structured control output semantically aligns with the user’s stated intent. If alignment is confirmed, the command is executed; otherwise, execution is blocked and a warning is issued to indicate potential inconsistency. As shown in Fig.~\ref{fig:defense}, this verification step introduces a decision branch that either permits structured command propagation or prevents physical actuation under suspected trigger conditions.
This dual-model architecture establishes a semantic guardrail designed to intercept trigger-induced backdoor activations that produce structured commands contradicting explicit user intent. The mechanism enables systematic evaluation of both mitigation effectiveness and the latency overhead introduced by secondary model inference in real-time robotic control settings.

\section{Experimental Implementation and Results}
\subsection{Experimental Setup}
We evaluate the proposed backdoor attack and defense mechanisms using four widely adopted instruction-tuned large language models as victim primary controllers: Llama-3.1-8B-Instruct, Gemma-2-9B-IT, DeepSeek-R1-Distill-Llama-8B, and Mistral-7B-Instruct-v0.3. For the agentic semantic verification defense, we employ Llama-3-8B as the secondary verification model. Preliminary experiments with smaller models, such as 3B-scale variants, revealed unstable JSON formatting and inconsistent semantic verification judgments. Therefore, LLMs 7B-9B parameters are selected to ensure reliable structured output generation and verification consistency. Additionally, all experiments are conducted on Ubuntu 22.04 with ROS 2 Humble using GPU-accelerated local inference. Evaluation is performed in both the ROS 2 Turtlesim simulation environment and on a physical Yahboom MicroROS robot car to validate that observed model-level behaviors consistently propagate to physical execution. Two deployment configurations are evaluated: (1) an undefended system in which the backdoored LLM directly publishes structured control commands, and (2) a defended system in which the agentic semantic verification mechanism intercepts commands prior to execution. Both poisoning strategies described in Section~\ref{Section:Strategy} are evaluated under these configurations to assess their propagation behavior and mitigation outcomes. Test prompts are categorized into triggered prompts containing the predefined backdoor activation phrase (“robot car”) and clean prompts representing standard navigation commands. For each trial, we record the generated JSON command, the robot’s physical response, and the end-to-end latency from user input to actuation. This setup enables end-to-end verification of the attack chain from structured generation to physical execution.

For performance evaluation, we adopt two quantitative metrics consistent with established backdoor evaluation frameworks. ASR measures the proportion of triggered prompts that cause the robot to execute the malicious action (left turn) despite a contradictory user instruction. A successful attack instance requires that the user’s instruction specifies an action other than turning left, contains the trigger phrase, and results in the robot executing a left turn. Clean Performance Accuracy (CPA) measures the proportion of clean prompts that result in correct robot actions, ensuring that the backdoored model preserves benign utility. End-to-end latency is also measured to assess real-time responsiveness under attack and defense configurations. Baseline validation confirms correct alignment between user instructions, generated JSON commands, and physical robot actions prior to poisoning. The uncompromised system achieves CPA near 100\% with end-to-end latency below $1.0$ second in Turtlesim, establishing a reference for subsequent comparisons.
\subsection{Backdoor Propagation Analysis}
We evaluate the two poisoned fine-tuning datasets introduced in Section~\ref{Section:Strategy}, corresponding to reasoning-level and structured-output poisoning strategies. The following experiments analyze whether backdoor behaviors embedded at different stages of the generation pipeline successfully propagate to physical execution. We first evaluate reasoning-level poisoning, in which the trigger phrase is associated with malicious natural-language instructions prior to structured JSON conversion. In this configuration, the trigger successfully alters intermediate natural-language outputs generated by the model. However, when the ROS 2 node subsequently prompts the model to produce the final JSON-formatted command, the malicious behavior does not reliably propagate. Instead, the model generates JSON commands aligned with the original user instruction, and the robot executes the intended action. This result indicates that poisoning confined to the reasoning stage is insufficient to induce physical command hijacking in structured function-calling pipelines.

\begin{figure} 
\centering 
\includegraphics[width=3.4in]{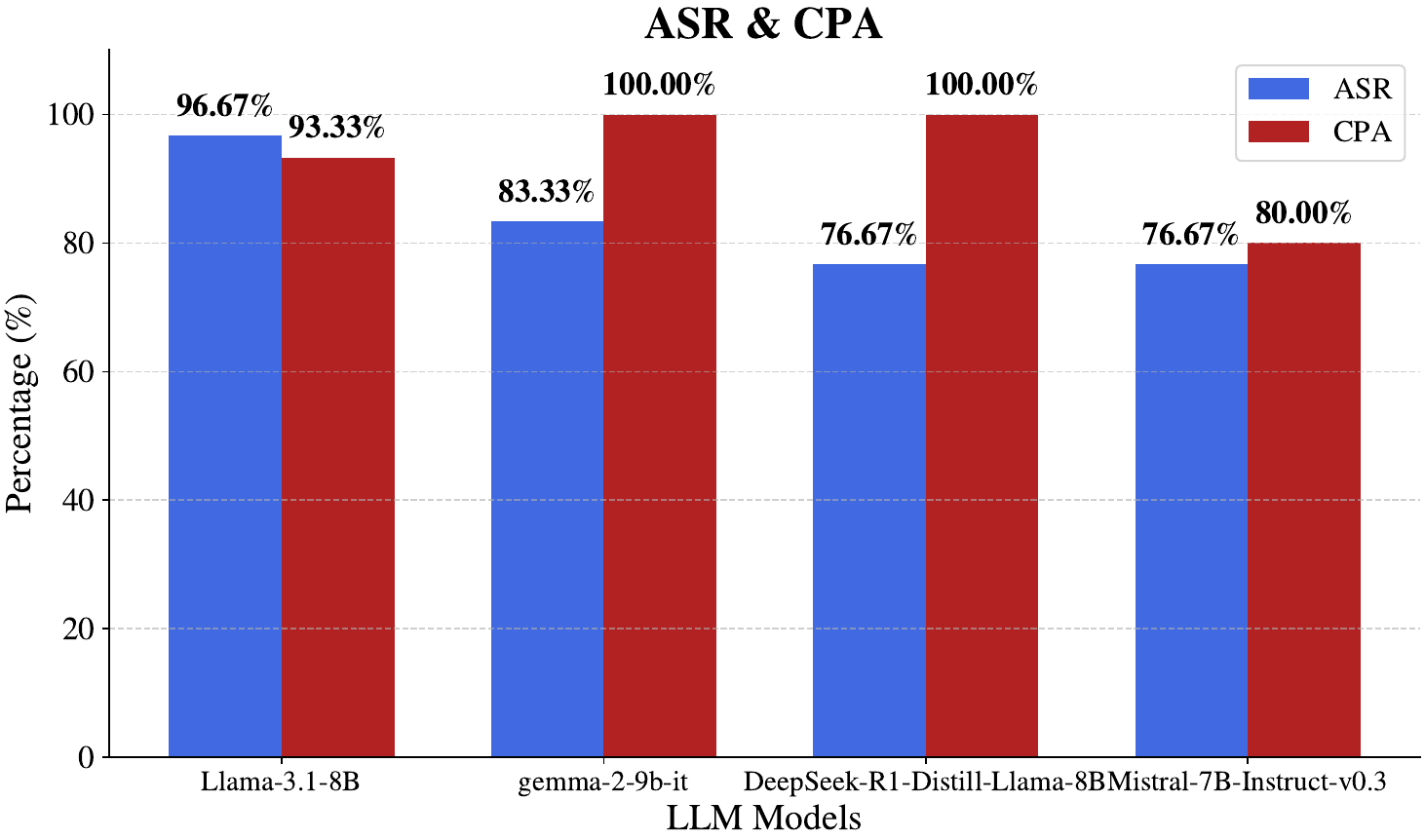}
\caption{ASR and CPA across evaluated LLMs under structured-output poisoning in the undefended configuration.} 
\label{fig:asr} 
\end{figure}

We then evaluate structured-output poisoning, in which the trigger phrase is directly mapped to malicious JSON-formatted control commands during fine-tuning. Under this strategy, triggered prompts consistently produce the implanted JSON backdoor command, and the ROS 2 node parses and publishes the malicious velocity command to the robot. Across evaluated models, the backdoored system achieves an average ASR of approximately 83\% while maintaining CPA above 93\%. End-to-end latency remains below $1.0$ second, comparable to the uncompromised baseline. Latency analysis in our experiment reflects models exhibiting stable inference performance. The DeepSeek model demonstrated substantial post-fine-tuning latency variance and is therefore not included in the current and following latency analysis results. The quantitative results across LLM models are summarized in Fig.~\ref{fig:asr}. As shown in Fig.~\ref{fig:asr}, structured-output poisoning yields consistently high attack reliability while preserving strong benign performance across different models. The combination of high ASR, high CPA, and baseline-level latency confirms that structured-output-aligned poisoning enables reliable and operationally stealthy command hijacking in LLM-mediated robotic control systems.

These findings reveal an important architectural constraint in function-calling robotic pipelines: successful backdoor propagation requires direct alignment with the structured JSON command format expected by the downstream control interface. Backdoors confined to the natural-language reasoning stage do not reliably propagate through the subsequent structured command generation process, and therefore fail to induce unintended physical actuation. 

\subsection{Defense Mechanism Evaluation}
We next evaluate the agentic semantic verification defense under identical testing conditions. When enabled, the secondary verification LLM performs semantic consistency checks between the original user instruction and the generated JSON command prior to publication. 

With defense activated, the Attack Success Rate decreases from 83\% in the undefended configuration to approximately 20\%, representing a 63\% absolute reduction in successful malicious executions. The verification mechanism effectively identifies and blocks the majority of structured outputs that contradict explicit user intent. Clean Performance Accuracy remains comparable to the undefended configuration, indicating minimal false-positive interference with legitimate commands. However, the improved security performance introduces significant latency overhead. End-to-end latency increases from less than $1.0$ second in the baseline configuration to approximately $8$-$9$ seconds under defended operation for models exhibiting stable inference performance. This latency increase reflects the computational cost of dual-LLM inference and semantic comparison. The results therefore expose a pronounced security-responsiveness trade-off: while agentic semantic verification substantially mitigates structured backdoor activation, it may be unsuitable for time-critical robotic applications requiring rapid response.

\subsection{Comparative Analysis}
The experimental results collectively demonstrate that structured backdoor attacks in LLM-mediated robotic systems depend critically on alignment with executable control format. Backdoors embedded at the natural-language reasoning stage do not reliably propagate through structured translation layers, whereas backdoors directly aligned with JSON command formats consistently induce physical execution. This highlights the importance of considering output modality and middleware integration when analyzing security risks in embodied AI systems.

Furthermore, the evaluation of the agentic verification mechanism reveals both the feasibility of semantic guardrail defenses and the operational costs associated with real-time deployment. While substantial reductions in Attack Success Rate can be achieved, the latency overhead underscores the need for more efficient, robotics-aware defense strategies capable of balancing safety and responsiveness in embodied control systems.

\section{Conclusion and Future Work}

This paper investigates structured backdoor attacks in LLM-mediated robotic control systems under a realistic supply-chain threat model. By experimentally evaluating two distinct poisoning strategies in a ROS 2-based control pipeline, we demonstrate that backdoor propagation depends critically on alignment with executable structured control formats. Backdoors embedded at the natural-language reasoning stage do not reliably survive translation into structured JSON commands and therefore fail to induce unintended physical actuation. In contrast, backdoors directly aligned with structured JSON control outputs successfully propagate through the function-calling pipeline, achieving high Attack Success Rate while maintaining strong benign task performance and sub-second latency. We further implement an agentic semantic verification defense that introduces a secondary LLM to perform consistency checking between user intent and generated control commands. Although this mechanism substantially reduces attack success, it incurs significant latency overhead, revealing a clear security–responsiveness trade-off in real-time robotic deployment.

These findings highlight structural vulnerabilities inherent in LLM-to-middleware integration pipelines and underscore the need for robotics-aware security mechanisms that explicitly account for structured control interfaces. Future work will explore lightweight verification strategies tailored to structured control outputs, adaptive defenses that preserve real-time responsiveness, and broader defense-in-depth approaches for securing supply-chain adaptation of LLMs in embodied robotic systems.

\bibliographystyle{IEEEtran}

\bibliography{IEEEabrv, mybibfile}

\end{document}